\documentclass[conference]{IEEEtran}
\IEEEoverridecommandlockouts
\usepackage{cite}
\usepackage{amsmath,amssymb,amsfonts}
\usepackage{algorithmic}
\usepackage{graphicx}
\usepackage{textcomp}
\usepackage{xcolor}
\def\BibTeX{{\rm B\kern-.05em{\sc i\kern-.025em b}\kern-.08em
    T\kern-.1667em\lower.7ex\hbox{E}\kern-.125emX}}

\usepackage{booktabs}%
\usepackage{makecell}
\usepackage{url}
\usepackage{subcaption}

\usepackage{enumitem}

\begin{document}

\title{Learning Underwater Active\\Perception in Simulation\\
}

\author{\IEEEauthorblockN{Alexandre Cardaillac and Donald G.\,Dansereau}
\IEEEauthorblockA{\textit{Australian Centre For Robotics}, 
\textit{School of Aerospace, Mechanical and Mechatronic Engineering}, \\
\textit{University of Sydney}, Sydney, Australia}
}

\maketitle

\begin{abstract}
When employing underwater vehicles for the autonomous inspection of assets, it is crucial to consider and assess the water conditions. These conditions significantly impact visibility and directly affect robotic operations. Turbidity can jeopardise the mission by preventing accurate visual documentation of inspected structures. Previous works have introduced methods to adapt to turbidity and backscattering, however, they also include manoeuvring and setup constraints. We propose a simple yet efficient approach to enable high-quality image acquisition of assets in a broad range of water conditions. This active perception framework includes a multi-layer perceptron (MLP) trained to predict image quality given a distance to a target and artificial light intensity. We generate a large synthetic dataset that includes ten water types with varying levels of turbidity and backscattering. For this, we modified the modelling software Blender to better account for the underwater light propagation properties. We validated the approach in simulation and demonstrate significant improvements in visual coverage and image quality compared to traditional methods.
\end{abstract}

\section{Introduction}

The growing demand for inspection and monitoring of underwater assets, both natural and man-made, requires the development of cost- and risk-efficient technologies \cite{chen_survey_2023}. Underwater vehicles are often deployed to perform these tasks, reducing operational constraints and human risks. However, the performance of the mission depends on the skill of the operator in the loop in the case of a remotely operated vehicle (ROV), and depends on the quality of the mission plan in the case of an autonomous underwater vehicle (AUV). Because underwater visibility varies significantly with location, time of day, and weather, operators and autonomous systems fail to consistently report high-quality documentation of the underwater asset. Therefore, there is an important need to be able to understand and adapt to local water conditions.

Underwater imagery is often degraded due to several light effects caused by the underwater environment. They include absorption, scattering, and marine snow \cite{Mobley1994}:
\begin{description}
    \item [Absorption] Attenuates the light over distance, resulting in colour distortions and loss of contrast.
    \item [Scattering] Caused by the change in direction by the light, either backward or forward scattering depending on the reflection angle. This leads to haziness and blurriness in the image.
    \item [Marine snow] Suspended particulate matter, also referred to as non-algal particles. These macroscopic floating particles can appear in images and intensify the aforementioned effects.
\end{description}
These effects directly affect the underwater robotic operations. When an AUV is employed, it is difficult to know if the data collected will be usable, and typical inspection and monitoring operations can last up to several hours. If the collected data is of poor quality, the operation was a waste of time and money. Additionally, if the vehicle is relying on visual features, it could become a hazard when it is unable to properly observe the environment.

Methods are emerging to overcome these issues, including active perception, which is about maximising information gain according to the current perception of the environment \cite{chen_active_2011}. This is done by dynamically updating the control plan of the vehicle. Such methods can enable underwater vehicles to always remain in visual range, and ensure the collection of high-quality data throughout the operation.

Previous related work in underwater active perception aims to minimise the distance between the vehicle and targets, ensuring the vehicle is always in sight \cite{xanthidis_resivis_2024}. Although the method can consider multiple targets, visual coverage is also minimised, resulting in longer operations in the context of inspections. Image quality is considered in \cite{sheinin_next_2016}, as information gain is optimised, however, the camera and the light source must be independently movable, which is a major constraint.

When training a model based on synthetic data and considering image quality, realistic rendering must be done. Major rendering software such as Blender \cite{blender} or Unreal Engine \cite{unrealengine} works best at above-water rendering, as they include numerous simplifications for underwater rendering.

In this work, we propose to improve the autonomy of AUVs with high manoeuvring capabilities that possess an adjustable-intensity illumination source. For this, we develop a learning-based approach to active underwater perception leveraging realistic synthetic data generation for training. The approach aims at ensuring the collection of high-quality data while maximising the visual coverage at the same time. This is done by providing online guidance suggestions such as distance to target and manipulating the artificial light intensity. Additionally, we propose a calibration step to obtain relevant optical information about the water column, which is then used to tune the model's output. The contributions can be summarised as follows:

\begin{itemize}
    \item An upgrade to Blender to improve simulation of underwater imagery, including more accurate light behaviour in water and models of the oceans.
    \item A method for in-situ water column property estimation using a monocular camera and adjustable illumination.
    \item A framework providing online guidance suggestions to maintain high-quality data collection and maximise visual coverage in a broad range of water conditions.
\end{itemize}

These contributions can lead to more consistent and faster-paced inspections and monitoring of underwater assets with more valuable visual documentation. The project code and resources are available on our project page at \url{https://roboticimaging.org/Projects/ActiveUW/}.

\section{Related Work}

\subsection{Underwater Imagery Simulation}

The oceans are challenging environments that can be hard to access and control. Additionally, there is a large diversity of water properties and conditions that affect both how water behaves and how water looks. This makes it difficult to collect a large and representative set of images. For these reasons, simulation plays a key role in underwater robotics. It prevents unnecessary deployments and enables experimental evaluation and validation of methods in a safe and controlled environment. Furthermore, simulators can be used to generate datasets for downstream tasks such as the training and validation of models.

Numerous underwater robotics simulators exist \cite{Manhaes2016, zhang_dave_2022, potokar_holoocean_2022}, however, most of them focus on robotic functionalities such as odometry sensors and environmental effects such as wind and current, and do not provide a realistic visual representation of the underwater scene as captured by an imaging device. The others are often tailor-made for specific applications, not adaptable, are private, or only release publicly datasets \cite{Zwilgmeyer2021,Ganoni2018,tunon2019}. There are two main approaches to generating underwater images: learning-based \cite{li_watergan_2018,ueda_underwater_2019,wang_uwgan_2021}, and model-based \cite{desai_rsuigm_2024,hou_benchmarking_2020,desai_rendering_2021}. Although the results may appear realistic, they are not always physically correct, as they are often based on a single image without contextual information, or generated using approximated models. Modelling software such as Blender or Unreal Engine attempt to render realistic and physically accurate images of the environments through advanced lighting simulation. However, since they are mostly used for above-water applications, the physics of underwater light are neglected. Indeed, the simulated light transport in participating media is not appropriate for underwater environments. Firstly, because the volume scattering phase functions employed significantly deviate from reality. These functions describe the light behaviour in a volume, more precisely, the probability distribution of light scattered in specific directions, indicating how light bounces in a medium \cite{PHARR2017671}. Blender uses the Henyey-Greenstein function \cite{Henyey1941} and Unreal Engine the Schlick function \cite{blasi_rendering_1993}. This deviation largely impacts simulations, especially for small and large angles, resulting in an incorrect representation of forward and backscattering, and stronger contrast in the image than expected, which are major factors limiting image quality in underwater environments, motivating the importance of using an appropriate model. Secondly, because the exponential decay of light is a function of colour coefficients with bounds, preventing the replication of real-world environments, instead of actual absorption and scattering density coefficients.

In this work we propose to implement and integrate modified physics for underwater light propagation into Blender to make the rendered underwater imagery more physically accurate.

\subsection{Active Perception}

In robotics, active perception enables platforms to react and adapt to their environment based on the information they gathered along the way. It is important to consider when the robot has to interact with the environment or optimise mission objectives, such as information gain or coverage.

Active perception is often used as a guarantee to avoid collisions and occlusions \cite{rodriguez-teiles_vision-based_2014,cortes-perez_low-cost_2016,ganesan_robust_2016,wedpathak_active_2016} where detection of obstacles results in motion re-planning. This ensures to have throughout the mission collision-free paths, and therefore safety.

In the context of underwater inspection, active perception is also employed to ensure sensor coverage of the asset by guiding the vehicle around the inspection target \cite{hover_advanced_2012,hollinger_active_2013,palomeras_autonomous_2019,cardaillac_application_2023,cardaillac_rov-based_2024}. It is either done to update online a reference path based on sensor feedback so that the new path adopts the shape of the structure, or done through the analysis of what has been covered and how, so that the vehicle knows how to correct and improve the documentation.

However, active perception is rarely used as a way to enhance the visual quality of the data collected. A control strategy for net inspection in turbid water is proposed in \cite{Lee2022}. It is based on the mean gradient feature used as a way to describe how much can be seen. The vehicle adapts its distance to reach a target mean gradient value and enable downstream tasks. Although this method works well in a variety of lighting conditions, it requires prior knowledge of the environment to setup some of the parameters such as target feature value and light source. In \cite{sheinin_next_2016}, the authors approach the problem as a next best view problem, manipulating both camera and light source placements, and where the optimisation criterion is information gain. While this method effectively reduces backscatter and increases contrast in captured images, the required setup makes it hard to deploy in the field.

Understanding the water column first can significantly improve the active perception framework as the water properties can provide insights on how the environment is and can be perceived. This can be done by measuring inherent optical properties (IOPs) and apparent optical properties (AOPs) which are often estimated using high-end sensors such as spectrometers and hyperspectral sensors \cite{bongiorno_coregistered_2018, Lovas:23}. However, such sensors are not always available. Therefore, alternative methods to derive water properties were developed using a camera as the main sensor. Light propagation models for the image formation process are explored \cite{akkaynak_revised_2018, 6907416}, however, they require the scene depth as input, and need to be recomputed when there is a change in illumination.

In this article we propose a simple but effective active perception framework that provides online suggestions to the vehicle in order to obtain better imagery. A calibration pattern is first followed to derive properties of the water column by capturing images along the depth and at varying distances to a target while turning an artificial light source on and off. The model then optimises image contrast and visual coverage over time by suggesting distances to a target and light intensity changes.

\section{Overview}

\begin{figure}[t]
    \centering
    \includegraphics[width=1\linewidth]{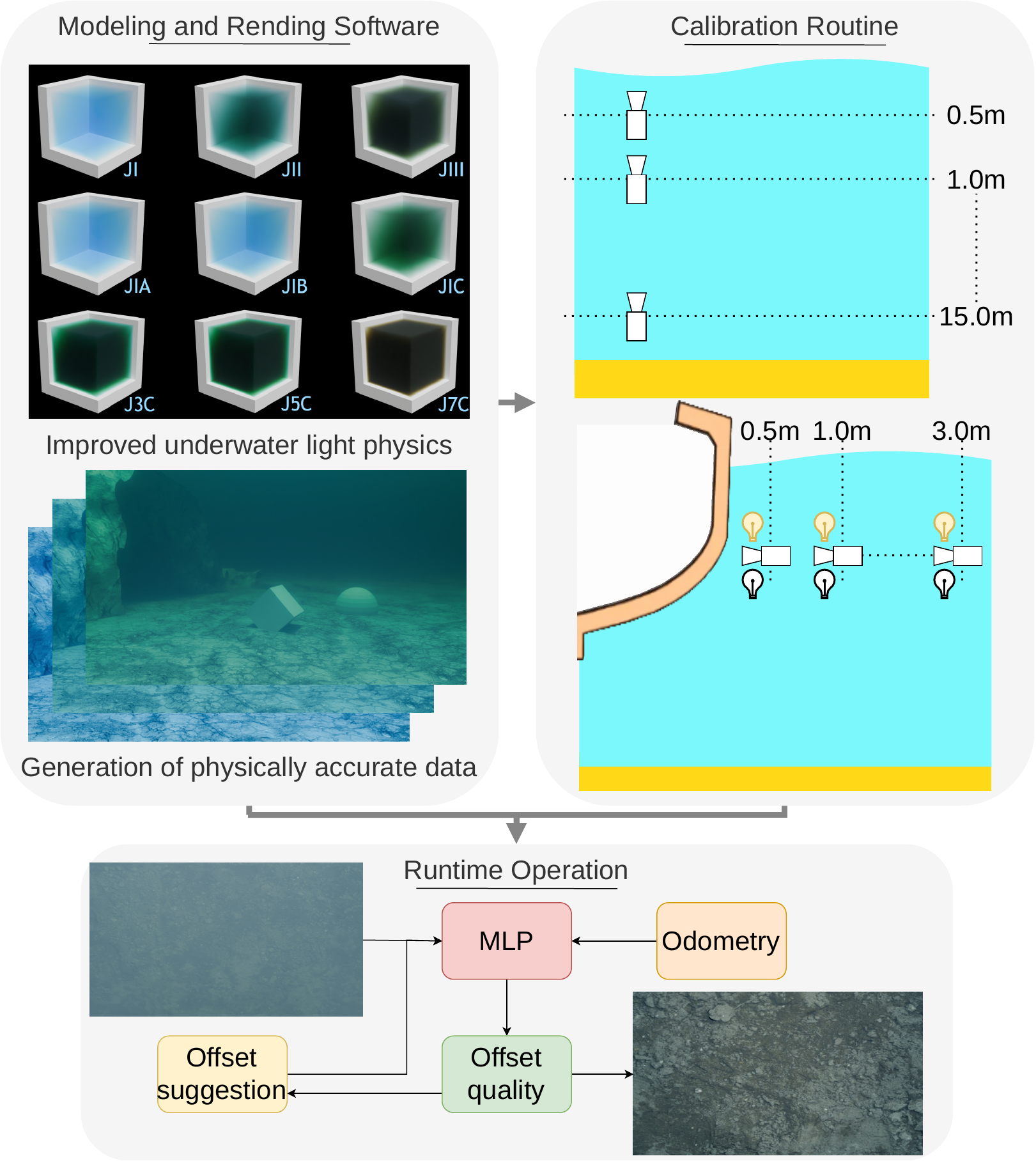}
    \caption{Overview of the proposed approach, divided into three components. We generate synthetic data (top left) in Blender including realistic physics and a large variety of water types. The vehicle follows the calibration routine (top right) to understand the water column and improve the runtime model (bottom) to better guide the vehicle and obtain better inspection data.}
    \label{fig:overview}
\end{figure}

The proposed work is composed of three main components: a modelling and rendering pipeline, a calibration routine, and a runtime model. An overview of the approach is depicted in Fig.~\ref{fig:overview}.

The first component is presented in Section~\ref{sec:simulation}. We first update the Blender software to better model light scattering and propagation underwater, allowing us to generate realistic and physically accurate data. We use the synthetic datasets for both, operation simulations and model training. This part is therefore enabling the experimentations and validations of the other two components.

The second component consists of a calibration routine: by manipulating the camera and manoeuvring the underwater vehicle, we can extract information about the water column, which describes the local visibility. This is discussed in Section~\ref{sec:calibration}. This component receives simulated images from the first component as the vehicle is moving and adapting its light settings.

The last component is about the runtime model, which proposes guidance suggestions to the vehicle to help collect better quality imagery. The calibration enables a prior knowledge of the environment while in mission, facilitating the runtime optimisation. This is explained in Section~\ref{sec:model}.

Finally, we evaluate the methods in Section~\ref{sec:experiments} using synthetic data. We demonstrate the applicability of the calibration routine and its importance, as well as how our approach compares to traditional methods.

In this project and throughout this paper, we use contrast as the image quality metric because it is influencing visual clarity and detail perception, and is largely impacted by the water column.

The contrast of an image is key in visual perception and computer vision operations, as it is influencing the quality of an image, including clarity and sharpness \cite{avatavului2023evaluating}. Contrast as a metric is used in a large variety of computer vision applications \cite{alma991032232771605106} and is essential for observing objects and features within an image as well are segmentation and recognition operations \cite{gonzalez2017,10.5555/574140,841534,10.5555/180895.180940}. This also holds true in underwater applications, where contrast is especially important for survey and inspection operations \cite{OByrneMichael2019IDAf}.

\section{Underwater Imagery Simulation}
\label{sec:simulation}

The underwater imagery simulation has multiple uses: on one hand, it helps generate large-scale data, and on the other hand, it helps for quick and efficient testing of the developed methods. For this, the rendered images must be as accurate as possible and reproduce the physical properties of the known oceans and seas.

We propose to update the Cycles rendering back-end of Blender to make it more suited for underwater light simulation. It is a physically and ray-trace-based production render engine that can realistically simulate participating media, including absorption, scattering, and emission. To adapt it to underwater environments, we replace the Henyey-Greenstein (HG) scattering phase function \cite{Henyey1941} with the Fournier-Forand (FF) function \cite{fournier_analytic_1994,Fournier1999}. The latter function better accounts for large and small scattering angles and the backscatter fraction, which are crucial for physically accurate underwater image rendering. The HG function was largely used in the past but has now been replaced by the oceanographic scientific communities with the FF phase function. It also includes more complicated but more realistic parameters. The FF phase function is defined as
\begin{multline}\label{eq:ff:1}
\tilde\beta_{FF}(\psi) = \frac{1}{4\pi\left(1-\delta_\psi\right)^2\delta_\psi^\nu}+(\nu(1-\delta_\psi)-(1-\delta_\psi^\nu)\\+\left(\delta_\psi(1-\delta_\psi^\nu)-\nu(1-\delta_\psi)\right)\text{sin}^{-2}\left(\frac{\psi}{2}\right))\\
+\frac{1-\delta_{180}^\nu}{16\pi\left(\delta_{180}-1\right)\delta_{180}^\nu}(3\text{cos}^2(\psi)-1),
\end{multline}
with
\begin{equation}\label{eq:ff:2}
\nu = \frac{3-\mu}{2},
\end{equation}
and
\begin{equation}\label{eq:ff:3}
\delta(\psi) = \frac{4}{3(n-1)^2}\text{sin}^2\left(\frac{\psi}{2}\right).
\end{equation}
$\mu$ is the slope parameter of the hyperbolic distribution, which describes the particle size distribution, and $n$ is the real index of refraction of the particles. $\nu$ and $\delta$ are used to simplify the equation, and $\delta_\psi$ is $\delta$ determined for the scattering angle $\psi$. When the phase function (\ref{eq:ff:1}) is integrated, it becomes possible to obtain the backscatter fraction

\begin{figure}[t]
   \centering\includegraphics[width=1\linewidth]{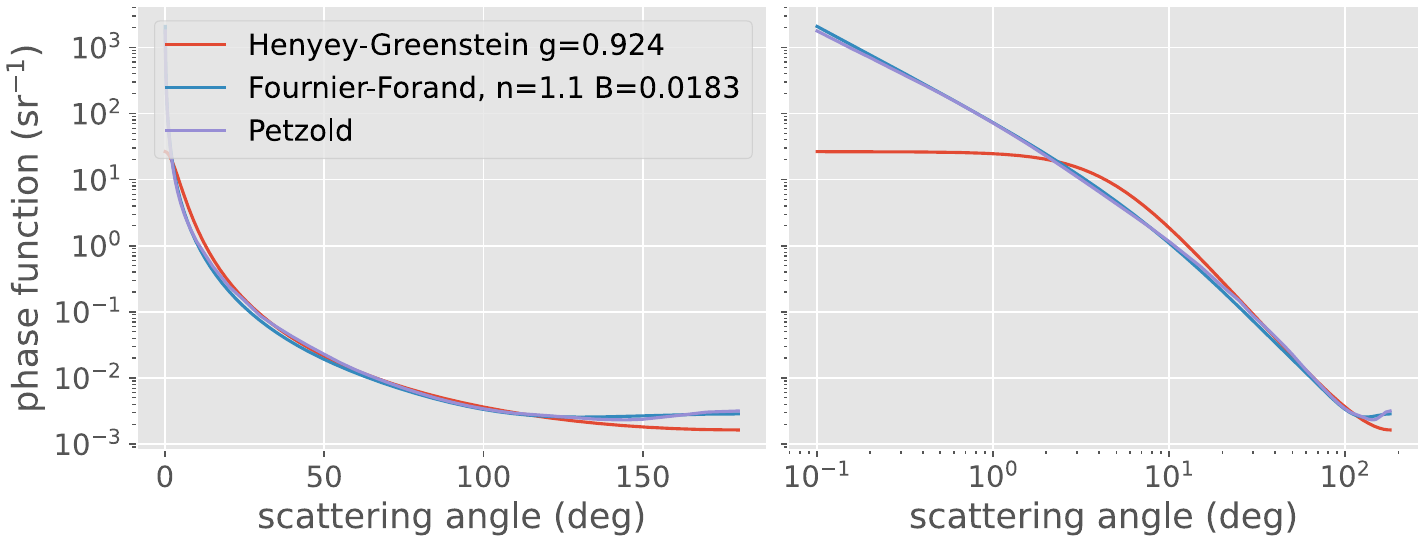}
    \caption{Comparison at different scales of the HG and FF phase functions, and the Petzold measurements. The HG function is far from the reference data, i.e., Petzold measurements, for low and large scattering angles, while the FF function closely matches the measurements.}
    \label{fig:pf_comp}
\end{figure}

\begin{equation}\label{eq:ff:4}
B = \frac{b_b}{b} = 1-\frac{1-\delta_{90}^{\nu+1}-0.5(1-\delta_{90}^{\nu})}{(1-\delta_{90})\delta_{90}^{\nu}}.
\end{equation}
However, in the context of rendering images, we are more interesting in controlling the backscatter fraction than the slope parameter of the distribution. Therefore, we rearranged equations (\ref{eq:ff:2}) and (\ref{eq:ff:4}) to express $\mu$ as a function of $B$,

\begin{equation}\label{eq:ff:5}
\mu = 2\frac{\text{log}\left(2B(\delta_{90}-1)+1\right)}{\text{log}(\delta_{90})}+3.
\end{equation}

\begin{figure}
    \centering
    \subfloat[]{
        \includegraphics[width=0.4\linewidth]{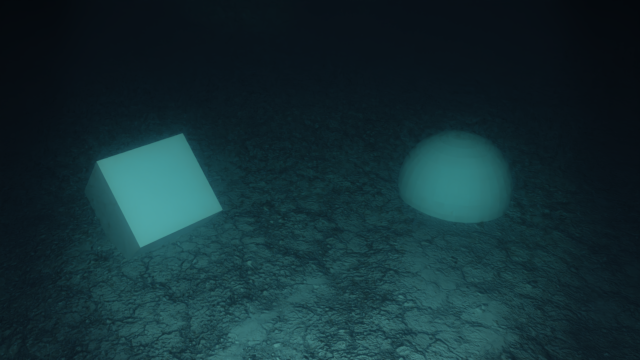}
        \label{fig:hg_ff:1}}
    \subfloat[]{
        \includegraphics[,width=0.4\linewidth]{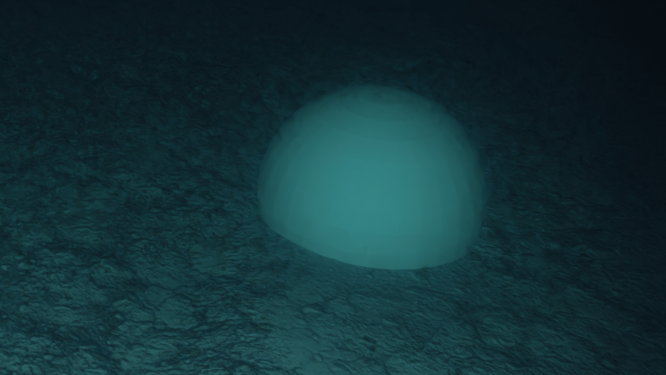}
        \label{fig:hg_ff:2}}\\
    \subfloat[]{
        \includegraphics[width=0.4\linewidth]{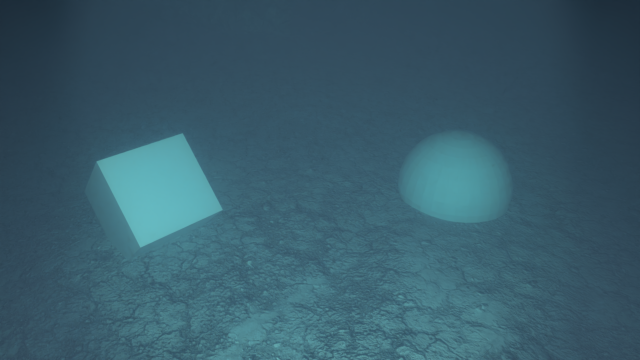}
        \label{fig:hg_ff:3}}
    \subfloat[]{
        \includegraphics[width=0.4\linewidth]{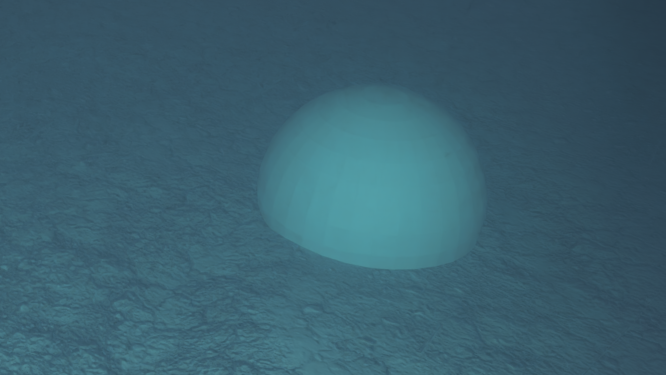}
        \label{fig:hg_ff:4}}
    \caption{The same scene is rendered using Blender but with two different phase functions: HG (top row) and FF (bottom row). The right column contains zoomed-in images. The bottom row images better represent backscatter and don't have as much contrast as the top row, which are known characteristics of underwater images.}
    \label{fig:hg_ff}
\end{figure}

Additionally, since Blender performs Monte Carlo based rendering, the corresponding sampling strategy and associated scattering angle determination must be updated. Since solving the probability density function of this phase function for an angle would lead to extensive computations, a binary search is achieved to find the angle corresponding to the given cumulative distribution function. To validate the implementation of equations (\ref{eq:ff:1}-\ref{eq:ff:5}), we compare the default phase function implemented in Blender, the HG function, to the FF function we added, and the Petzold measurements \cite{Petzold1972VolumeSF}, used as reference, corresponding to actual measurements in oceans. The Petzold measurements serve as the basis for understanding how light is scattered in different underwater environments and are still today the most carefully made and widely cited scattering measurements \cite{Mobley1994}. Fig.~\ref{fig:pf_comp} shows the comparison results. The differences between the two functions appear clearly for small and large scattering angles, while the FF function matches closely the reference data.

When rendering images with both phase functions, the visual difference is significant and entirely changes the contrast of the image, especially in environments with high scattering values and with underwater light sources. The backscattered light becomes naturally apparent using the FF function because it presents a higher probability of light scattering for large scattering angles, as it can be seen in the Blender renders displayed in Fig.~\ref{fig:hg_ff}. In this setup, we used the same values as in Fig.~\ref{fig:pf_comp}, and added a light source next to the camera pointing at the objects in the scene. Backscatter making the scene appear more hazy is expected. The figure shows the FF function is able to capture this effect, whereas the HG function is not.

\begin{figure}[b]
    \centering
    \subfloat[]{
        \includegraphics[width=0.32\linewidth]{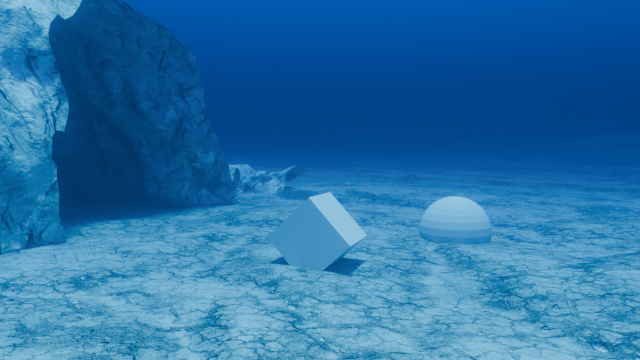}}
    \subfloat[]{
        \includegraphics[width=0.32\linewidth]{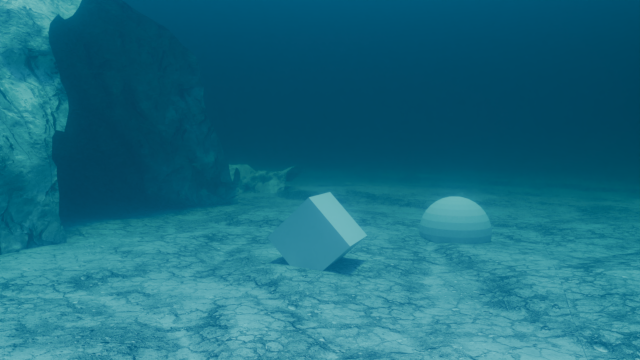}}
    \subfloat[]{
        \includegraphics[width=0.32\linewidth]{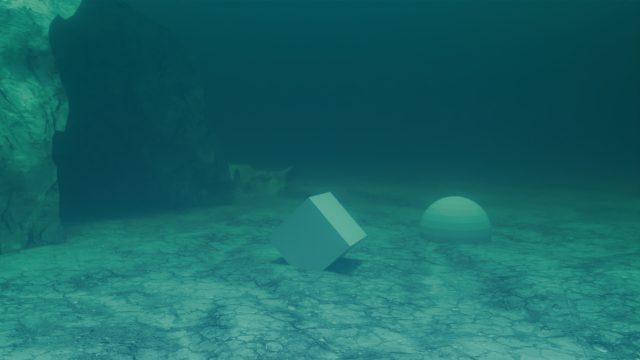}}
    \caption{Renders with similar absorption characteristics but the scattering coefficient is gradually increasing from left to right.}
    \label{fig:blender_renderings}
\end{figure}

We proposed this addition to the Blender development team which accepted it and included it as part of the official 4.3 release of the software, along with the integration of other phase functions. The implemented phase functions can now be selected from the front-end interface of the software. Example environments generated in Blender are presented in Fig.~\ref{fig:blender_renderings}.

\begin{figure}[t]
    \centering
    \includegraphics[width=0.6\linewidth]{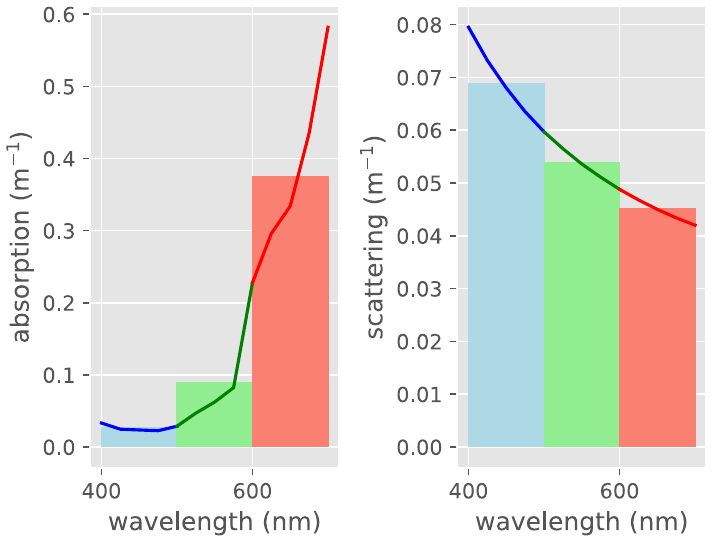}
    \caption{Example of the conversion from wavelength to wideband coefficients, here using the data of the Jerlov IB ocean type.}
    \label{fig:conversion}
\end{figure}

Furthermore, we propose an easy-to-use Blender front-end tool to create oceans of various types. This includes being able to easily modify the ocean light absorption, scattering, and colour coefficients. The major ocean types are included, based on the inherent optical properties of Jerlov water types \cite{jerlov1976marine,Solonenko2015} with generalisation of the wavelength-based coefficients into wideband coefficients. Ten water types are included, from coastal waters to oceanic waters, with various levels of turbidity. The conversion from wavelength to wideband coefficients consists of taking the average value of specific ranges, by default, $[600, 700]$ for red, $[500, 600]$ for green, and $[400, 500]$ for blue. The intervals can be changed to match specific camera characteristics. The case of the Jerlov IB water type is presented in Fig.~\ref{fig:conversion}. This front-end extension can be found on our project website.

Finally, we add physics-based noise to the rendering pipeline since Blender does not account for sensor noise. We employ the noise model from \cite{foi_clipped_2009}, which was developed to enhance the realism of simulated images, and it includes thermal and signal-dependent Poisson noise. 

\section{Water Column Property Estimation}
\label{sec:calibration}

To obtain better quality imagery while underwater and in operation, it is important to understand the water column properties. When using the camera as a photometric sensor, properties can be derived by observing the environment with different angles and illumination settings.

To observe the water column properties, we propose a calibration routine in two steps to perform at the beginning of the underwater mission. The first one involves observing the change in illumination over depth while looking towards the water surface, this provides insights into how is light attenuated as it penetrates the water column. The second step is about measuring the change in contrast over distance to an object with lights on and off, to better understand ambient illumination and backscattering properties of the environment.

Both distance and depth depend on the operation requirements. For example, an operation occurring at 5m depth and at a typical average distance of 2m from the target would need the calibration to occur at up to 6m depth and 3m distance. The calibration routine enables to estimate both the channel-wise colour attenuation over depth, and the channel-wise contrast change over distance. Visual representations of the routine are shown in Fig.~\ref{fig:overview}, middle column.

\begin{figure}[t]
    \centering
    \subfloat[]{
        \includegraphics[width=0.41\linewidth]{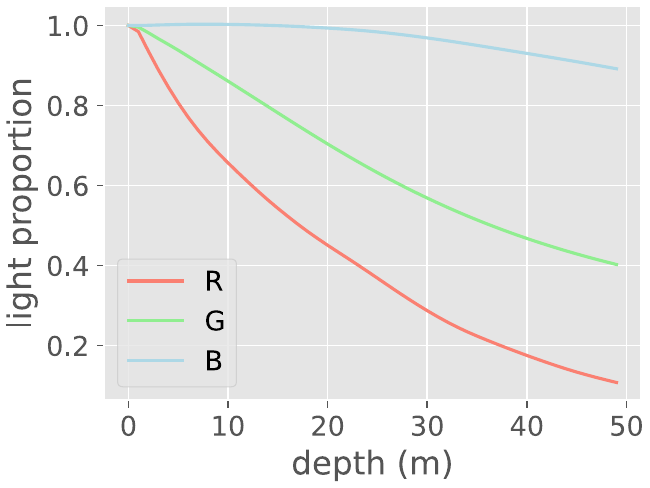}
        \label{fig:calib_values:1}}
    \subfloat[]{
        \includegraphics[width=0.41\linewidth]{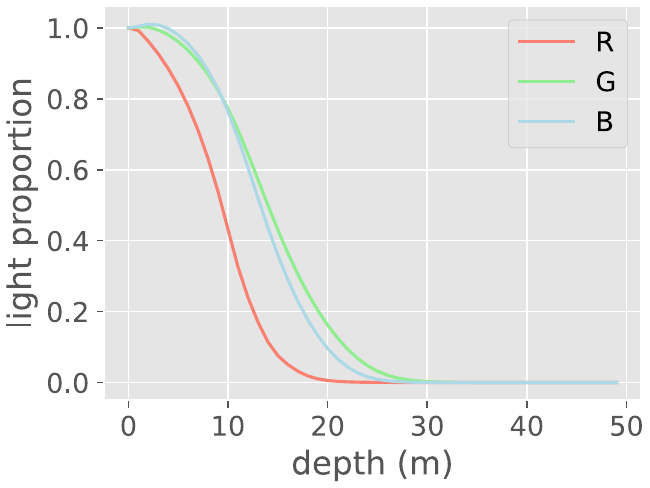}
        \label{fig:calib_values:2}}\\
    \subfloat[]{
        \includegraphics[width=0.41\linewidth]{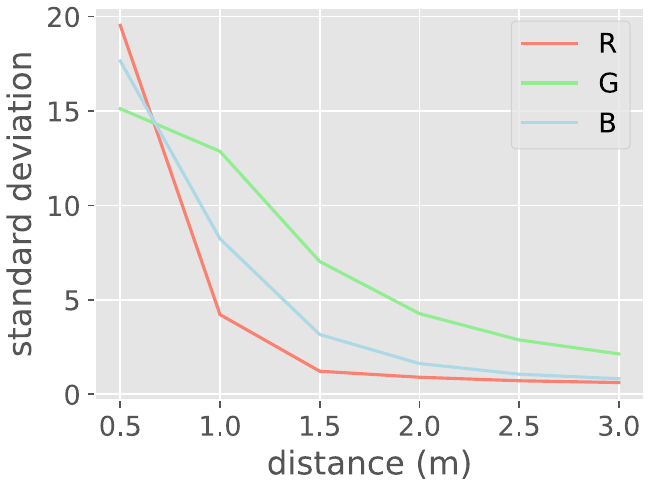}
        \label{fig:calib_values:3}}
    \subfloat[]{
        \includegraphics[width=0.41\linewidth]{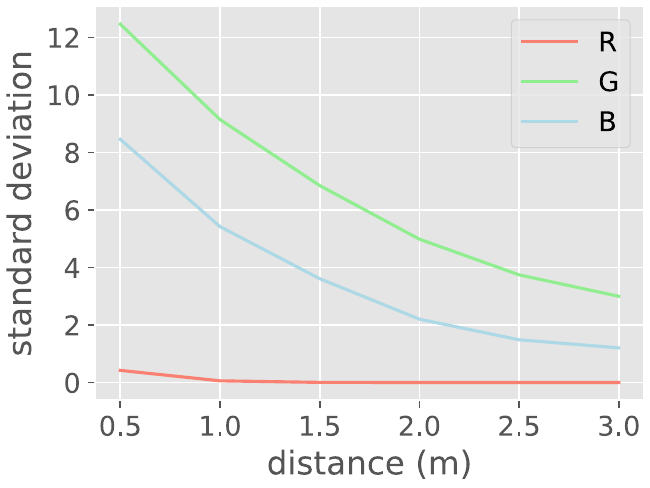}
        \label{fig:calib_values:4}}
    \caption{Example calibration profiles where the first row includes two examples of depth profiles, \ref{fig:calib_values:1} the JI water type with a backscatter coefficient of 0.005, and \ref{fig:calib_values:2} the JII water type with a backscatter coefficient of 0.0183. The second row includes the contrast profiles of the water type J3C with 0.005 backscatter, \ref{fig:calib_values:3} with lights on, and \ref{fig:calib_values:4} lights off.}
    \label{fig:calib_values}
\end{figure}

The colour attenuation over depth is estimated as a ratio, corresponding to the proportion of the total local illumination, using as reference the closest measurement to the water surface, such that
\begin{equation}
K_z(z) = \frac{\mu(z)}{\mu(0)},
\end{equation}
where $z$ is the depth, and $\mu(z)$ is the mean of the image taken at depth $z$. The second parameter, the contrast over distance, is simply measured using the standard deviation of the images. To estimate this, each image collected along the distance is divided into patches $p$ with standard deviation $\sigma_p$. The average standard deviation $\Bar{\sigma}_p$ is computed by taking the channel-wise average of all the patch standard deviations. The contrast over distance is then given by
\begin{equation}
K_c(d) = \Bar{\sigma}_p(d),
\end{equation}
where $d$ is the distance, and $\Bar{\sigma}_p(d)$ is the patch standard deviation of the image taken at distance $d$. This operation is repeated two times, once with the lights on, and once with the lights off, providing insights on the light scattering. Using patches enables to extract more relevant image statistics, especially in the presence of contrast distributed over the image which can create biases. The contrast of an object decreases over distance while underwater. Simply measuring the standard deviation of the captured image might not reflect this underwater property.

When accumulating these measurements over depth and distance, we obtain channel-wise profiles that serve as reference values during the underwater mission. However, the operation must remain in the depth and distance bounds defined during the calibration. Examples of profiles are presented in Fig.~\ref{fig:calib_values}. They include both profiles, contrast over distance and light attenuation over depth. These profiles are then used as reference data during the operation. Depending on the operation requirement, the sampling distance might vary. For example, in both Fig.~\ref{fig:calib_values:3} and Fig.~\ref{fig:calib_values:4}, estimations are computed every half meter, and provide accurate enough profile for the operation.

\section{Perception Aware Distance and Illumination Suggestions}
\label{sec:model}

\begin{figure}[t]
  \centering\includegraphics[width=0.65\linewidth]{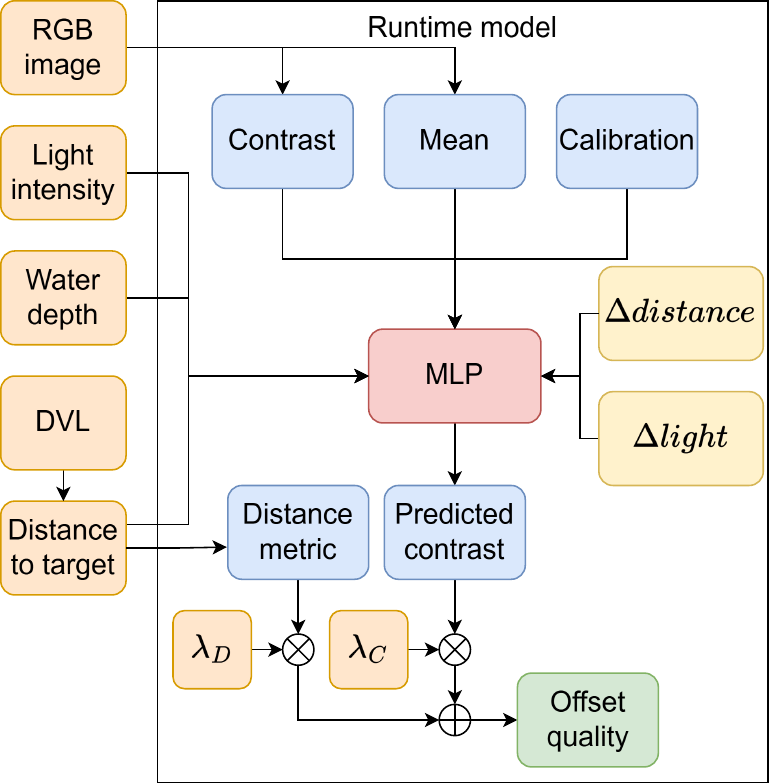}
    \caption{The runtime model provides guidance suggestions based on the MLP which predicts image contrast given a change in illumination and distance.}
    \label{fig:quality_pred}
\end{figure}

Adjusting the trajectory of the underwater platform and its artificial illumination in real-time allows it to adapt to local conditions, including turbidity and illumination changes. This ensures the collection of high-quality imagery throughout the operation. To achieve this, we propose a framework providing guidance suggestions to the vehicle. They include distance and illumination changes. The best pair is found through an optimisation loop balancing image quality and coverage. Fig.~\ref{fig:quality_pred} summarises the main components of the proposed work. Here, we measure the distance using the acoustic beams of a Doppler velocity log (DVL) as it is a sensor that is regularly found on underwater vehicles. It could be replaced by other sensors such as a laser or a pinger.

\begin{figure}
        \centering\includegraphics[width=0.45\linewidth]{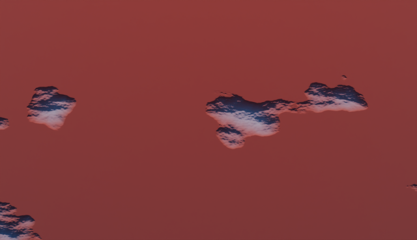}
        \centering\includegraphics[width=0.45\linewidth]{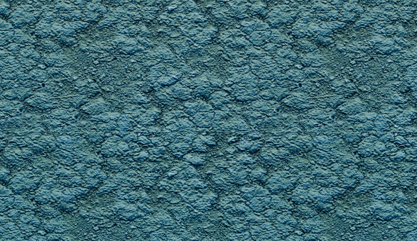}\vspace{0.1cm}
        \centering\includegraphics[width=0.45\linewidth]{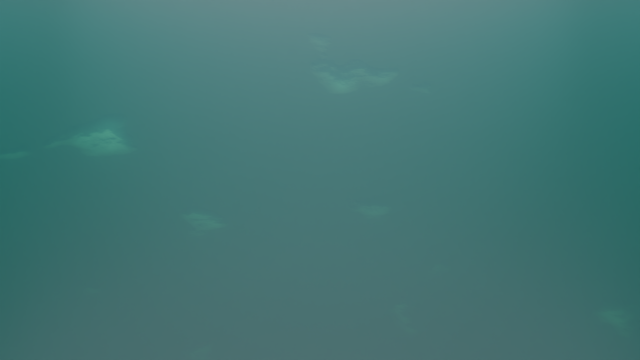}
        \centering\includegraphics[width=0.45\linewidth]{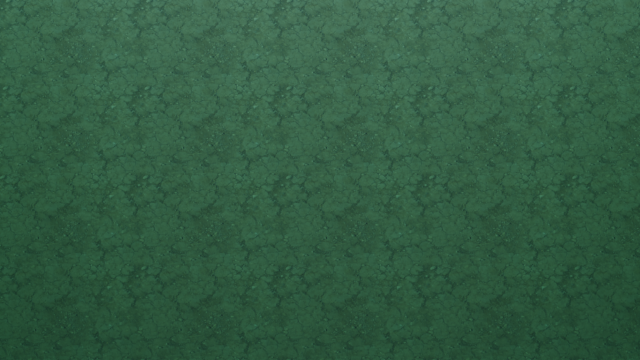}
        \caption{The two textures included in the dataset are a low-feature red ship hull with apparent paint peel (left), and a rich and highly detailed seabed (right). The images are captured in clear water (top row), and turbid water (bottom row).}
    \label{fig:envs}
\end{figure}

The first step is to be able to predict the image quality given a set of settings and a proposed pair of distance and illumination offsets. In this work, we use contrast as the quality metric. To enable this prediction step, we train a multi-layer perceptron (MLP) on a synthetic dataset generated with Blender. A total of 138240 entries were generated and includes: two model textures, ten water types, four backscatter coefficients, three depths, six distances to target, four light intensities, six delta distances, and four delta light intensities. Fig.~\ref{fig:envs} depicts the two textures used to generate the dataset. They were chosen based on the project objectives and to include diversity in the data.

The MLP has 20 inputs. Nine of them are from the calibration and are channel-wise parameters:
\begin{itemize}
    \item Global illumination change over depth (3)
    \item Contrast change over distance with lights on (3)
    \item Contrast change over distance with lights off (3)
\end{itemize}
Nine more inputs are from the current state of the operation:
\begin{itemize}
    \item Distance to target (1)
    \item Active illumination intensity (1)
    \item Current depth (1)
    \item Channel-wise mean of the current camera image (3)
    \item Channel-wise standard deviation of the current camera image (3)
\end{itemize}
Finally, the last two inputs are the parameters to optimize: the distance offset and the light intensity offset.

There are three outputs and correspond to the predicted channel-wise contrast of the image given the offset parameters. However, using this model alone will always favour getting the vehicle closer to the target, as contrast always decreases with distance underwater. Additionally, the closer the vehicle is, the smaller the coverage area, and hence, the longer it takes to survey the asset. Finding the right balance between image contrast and coverage is essential. Therefore, we add a distance regulator, and we use the weighted sum of the regulator and predicted contrast as the final output to obtain the following offset quality:
\begin{equation}
    y = \lambda_c m_c + \lambda_d m_d.
\end{equation}
$m_c$ and $m_d$ are respectively the predicted contrast and distance metrics, and $\lambda_c$ and $\lambda_d$ are their respective weights. The metric for the predictive contrast is simply defined as the sum of the three channels, while the distance metric is defined as
\begin{equation}
m_d = \exp\left(-\frac{(d-b)^2}{2c^2}\right),
\end{equation}
\begin{equation}
b = \frac{\sqrt{\displaystyle\sum_{R,G,B}{K_c(d)}}\,\kappa_b}{z} \qquad c = \frac{\Bar{K_z}(z)^2}{\kappa_c}.
\end{equation}
$b$ is defined according to the contrast profile, and $c$ according to the depth profile, where $\kappa_b$ and $\kappa_c$ are tuning parameters enabling prioritisation of the profiles.

To find the best pair of offset parameters $\Delta d$ and $\Delta l$, we integrated the MLP within an optimisation loop that maximises the quality output $y$. We found empirically that the Nelder-Mead optimisation method\cite{Nelder_1965} performed best. Once the best value of $y$ is found, the two offsets are saved and sent to the guidance system of the vehicle so that it can collect better imagery.

\section{Experiments}
\label{sec:experiments}

In this section, we validate the approach and methods in simulation using realistic and physically accurate data. We aim to first verify the patch-based image statistics extraction, followed by the performance evaluation of the model and metrics, and finally, we simulate an underwater operation and compare it to traditional and naive approaches.

\subsection{Image Characteristics}\label{sec:img_charac}

\begin{figure}
    \centering
    \subfloat[]{
        \includegraphics[width=0.45\linewidth]{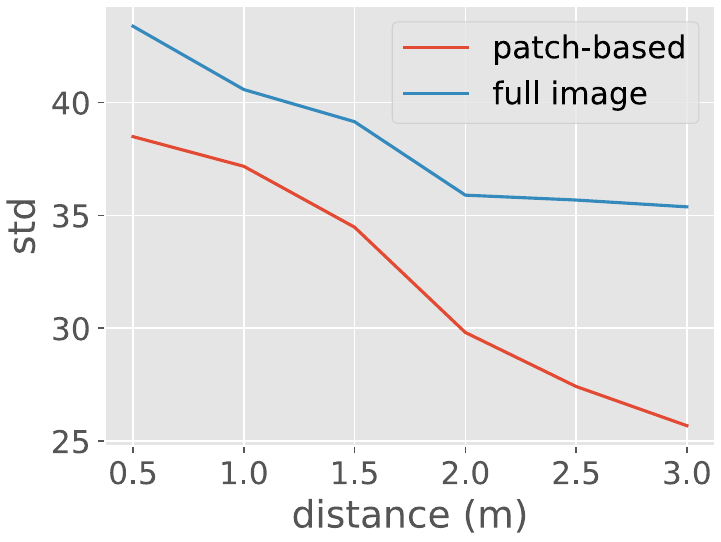}
        \label{fig:img_std:1}}
    \subfloat[]{
        \includegraphics[width=0.45\linewidth]{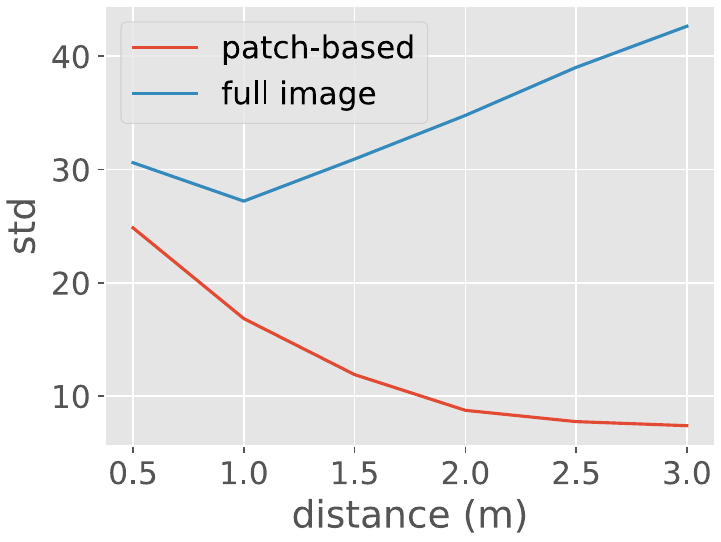}
        \label{fig:img_std:2}}
    \caption{Standard deviation comparison with image and patch-based operations in low turbidity \ref{fig:img_std:1} and high turbidity \ref{fig:img_std:2}. The patch-based method replicates more accurately the expected contrast change over distance while underwater.}
    \label{fig:img_std}
\end{figure}

To obtain more realistic statistics from the images, patch-based operations are performed instead of using the entire image. Here we validate that contrast decreases over distance. Each image is divided into 60 patches identical in size. For an image of size 1280x720 px, patches of 128x120 px are extracted, fitting 10 of them horizontally, and 6 vertically. Patches need to be small enough to remove the biases but also large enough to properly capture contrast. The comparison is displayed in Fig.~\ref{fig:img_std} using highly textured images with high frequency content. The differences are significant, especially in an environment with a high backscatter coefficient (\ref{fig:img_std:2}) where the contrast increases over distance when using the entire image due to the ambient illumination. At 3 meters distance, the target features are hard to observe due to the lack of contrast, which would be contradicted if patch-based operations were not performed. In low turbidity, the difference is not as important because at short distances, the target remains easily observable.

\subsection{Model Evaluation}

We built the MLP used for the contrast prediction as a TensorFlow \cite{tensorflow2015-whitepaper} model, and includes two hidden dense layers of 128 nodes, 20 inputs, and 3 outputs. We used the dataset introduced in Section~\ref{sec:model} for the training and testing of the model. A 20\% split was applied, and the model was trained over 25 epochs using the mean absolute error (MAE) as the loss function. For comparison, we trained a similar model, but without the calibration parameters. The results are shown in Table~\ref{tab:models}. On the test set, the standard model consistently achieved around 0.01 MAE, while the model without the calibration reached 0.05 MAE. We further tested the two models in two new environments, low turbidity and high turbidity environments. Similarly to the test set, the calibrated model is performing better in both settings. The difference is more significant in the low turbidity environment because lower variations of contrast are observed over distance and depth, which can be difficult to capture without a reference, i.e., the calibration parameters.

\begin{figure}
    \centering
    \subfloat[]{
        \includegraphics[width=0.8\linewidth]{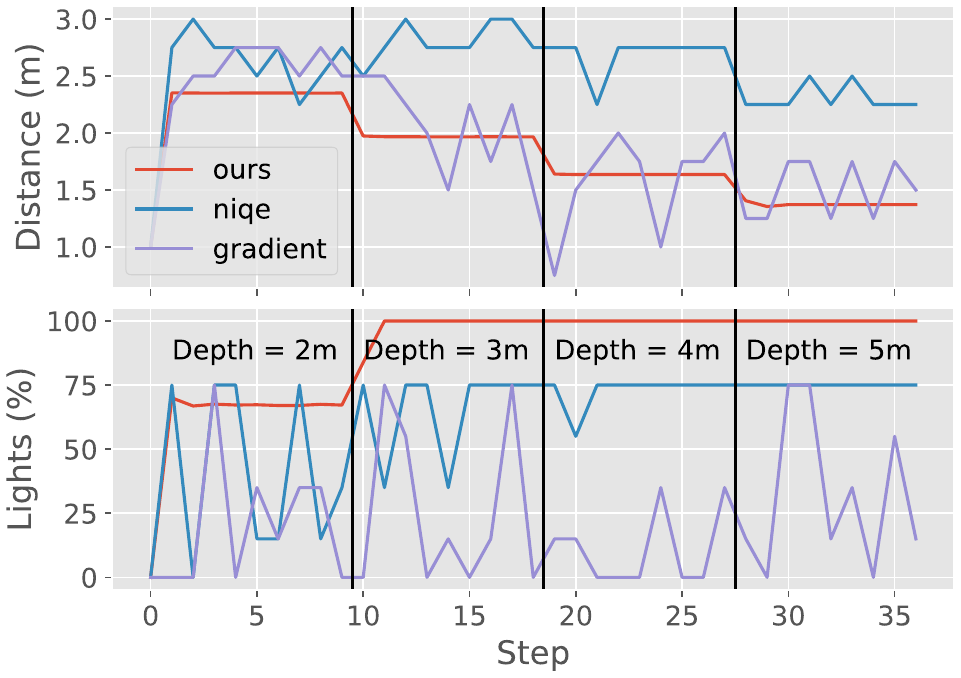}
        \label{fig:operation:1}}\\
    \subfloat[]{
        \includegraphics[width=0.8\linewidth]{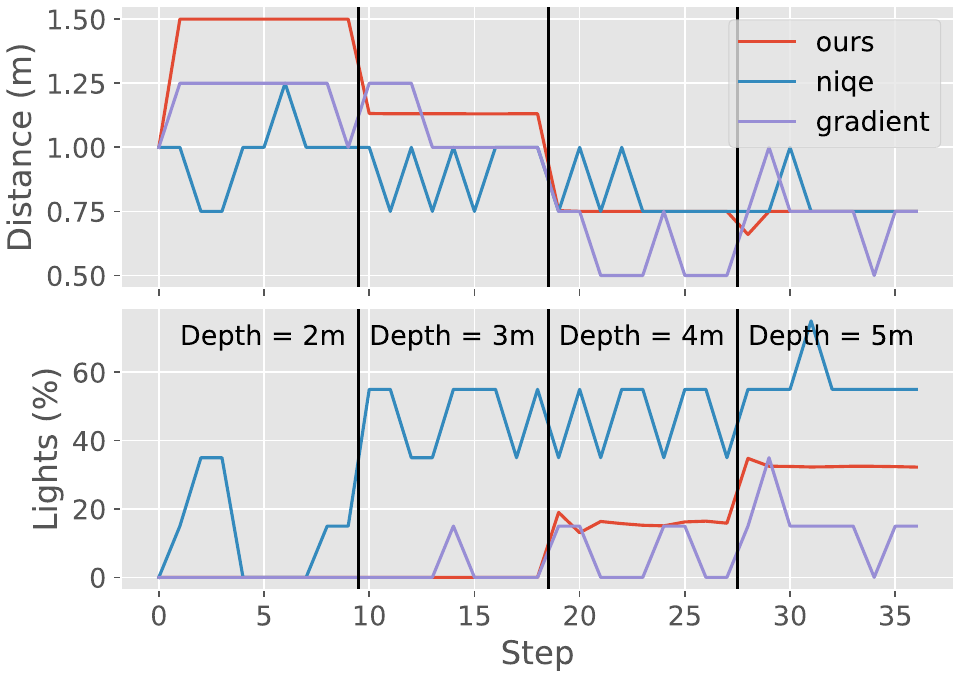}
        \label{fig:operation:2}}
    \caption{The optimisation over time of the distance and illumination of different methods is compared in low \ref{fig:operation:1} and high \ref{fig:operation:2} turbidity environments. Each approach adapts differently according to depth and turbidity.}
    \label{fig:operation}
\end{figure}

\begin{table}[b]
\caption{MAE of the contrast prediction\label{tab:models}}
\centering
\begin{tabular}{cccc}
\toprule
    Model                        & Test Set & Low Turbidity & High Turbidity \\ \midrule
    MLP w/ calibration  & \textbf{0.01}              & \textbf{0.09}                   & \textbf{0.04}                    \\
    MLP w/o calibration & 0.05                       & 0.49                            & 0.15                             \\ 
    \bottomrule
\end{tabular}
\end{table}

\subsection{Operation Evaluation}

To ensure the benefits of the proposed approach, a typical underwater inspection is simulated in two different water types: one with low turbidity, and one with high turbidity. For this, images are rendered along a lawnmower pattern. Every 50cm a set of images is rendered, with distance to target varying from 0.5m to 3m, and light intensity varying from 0\% to 100\%. During the evaluation, for each position, the image with the closest parameters to the prediction of the model is selected. This way, a path is generated and can be compared to more classical and static approaches, i.e., patterns at fixed distances and with fixed light intensity. Here, we chose to compare against patterns at 1 and 2 meters, and 50\% and 25\% light intensity respectively. On top of the classical methods, we are comparing against a gradient-based method inspired from \cite{Lee2022} where the objective is to maintain constant a distinctive mean gradient of the image after applying the Sobel filter \cite{sobel} in the y-direction. We also compare with a similar method, but using the natural image quality evaluator (NIQE) metric \cite{6353522} instead.

\begin{figure}
    \centering
    \subfloat[]{
        \includegraphics[width=1\linewidth]{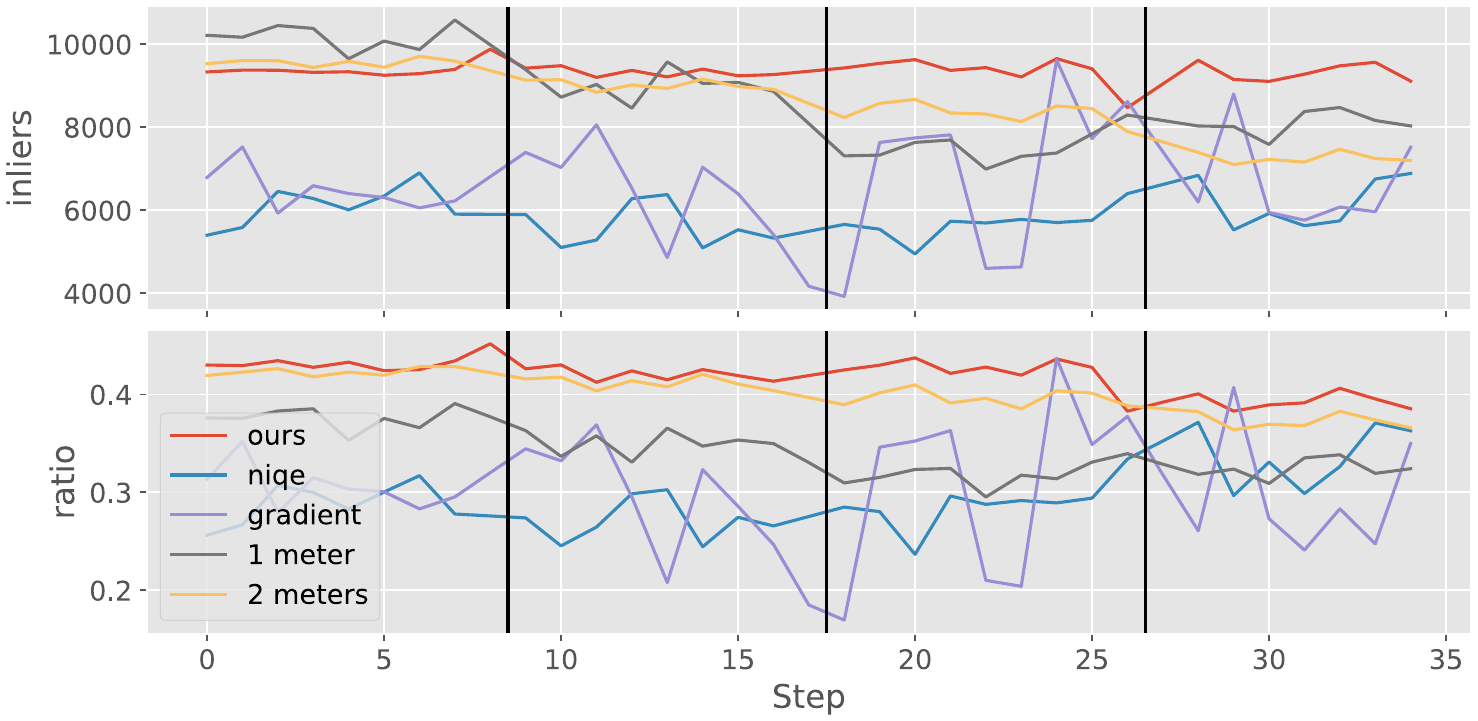}
        \label{fig:inliers:1}}\\
    \subfloat[]{
        \includegraphics[width=1\linewidth]{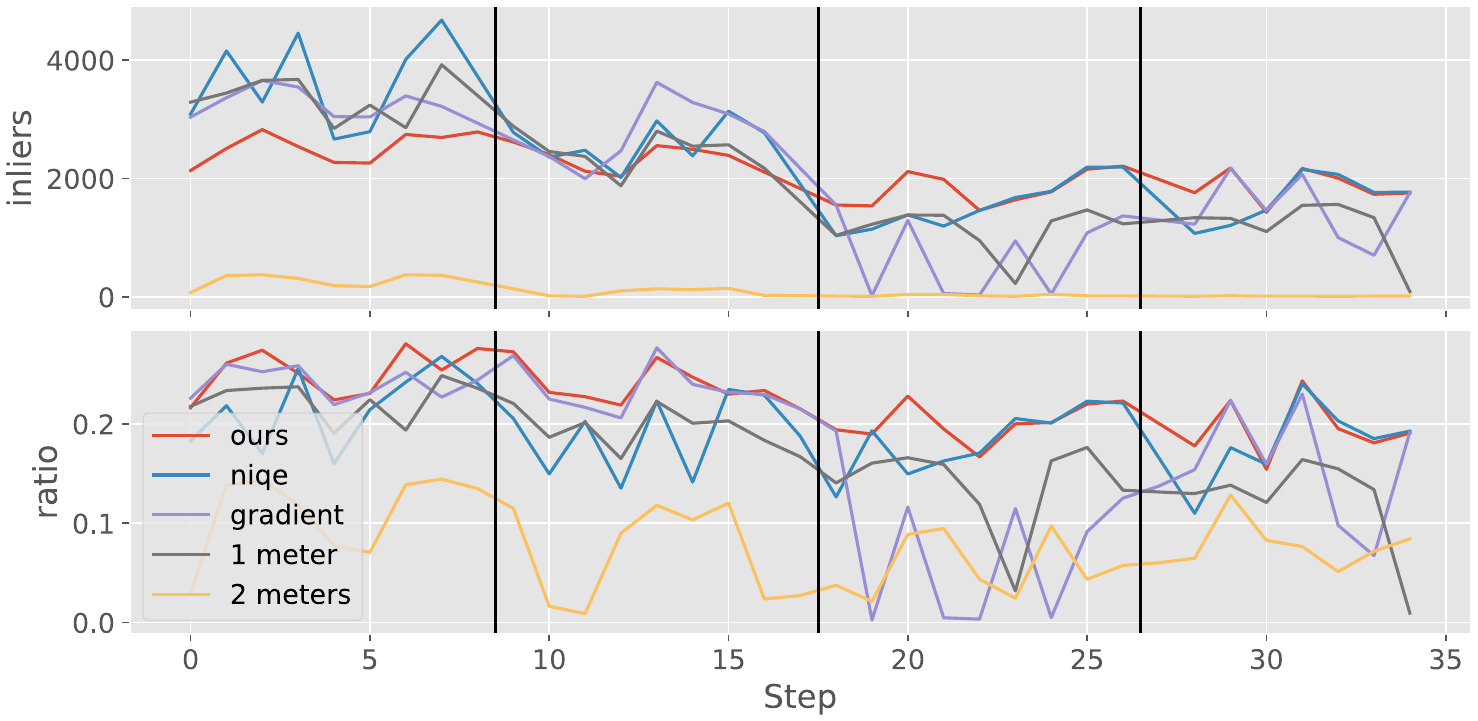}
        \label{fig:inliers:2}}
    \caption{The number of feature matching inliers and inlier ratios are compared in two environments, low \ref{fig:inliers:1} and high \ref{fig:inliers:2} turbidity. The proposed method shows more consistency and higher numbers on average.}
    \label{fig:inliers}
\end{figure}

The evaluation is done in three steps. First, a qualitative evaluation is done by observing how the parameters change over time (Fig.~\ref{fig:operation}). Secondly, evaluating the mapping capabilities by measuring how well the scene can be tracked (Fig.~\ref{fig:inliers} and Table~\ref{tab:inlier}). And finally, the total visual coverage is estimated (Table~\ref{tab:coverage}).

\begin{table}[t]
\caption{Comparison of averaged feature statistics in different turbidity levels\label{tab:inlier}}
\centering
\begin{tabular}{ccccccc}
\toprule
& \multicolumn{3}{c}{Low turbidity} & \multicolumn{3}{c}{High turbidity}                       \\ 
\cmidrule(r){2-4}\cmidrule(l){5-7}
Setup        & Features & Inliers & Ratio  & Features & Inliers & Ratio  \\ \midrule
Proposed     & 22391    & \textbf{9363}    & \textbf{0.42}   & 9527     & 2125    & \textbf{0.22}   \\
NIQE         & 20067    & 5904    & 0.30   & \textbf{12097}    & \textbf{2302}    & 0.19   \\
Gradient     & 22051    & 6620    & 0.30   & 11122    & 1999    & 0.17   \\
1m pattern   & \textbf{25290}    & 8670    & 0.34   & 11261    & 2087    & 0.18   \\
2m pattern   & 21362    & 8594    & 0.40   & 1079     & 112     & 0.08   \\
\bottomrule
\end{tabular}
\end{table}

The optimised parameters are tracked over time in Fig.~\ref{fig:operation} with the vehicle following the pattern. In the low turbidity environment, the vehicle is expected to stay at a long range from the target as there is clear visibility. This also makes it possible to turn on the lights and increase their intensities. However, in very turbid waters, the vehicle should remain close to the target and with low light intensity because of the risk of backscatter. The results reflect the expectations as each method follows the same pattern, i.e., getting closer and increasing the light intensity as the vehicle gets deeper, and at different rates for each method. However, only the proposed approach shows stability at each depth which is important for safe and efficient operations.

To ensure the proposed method also enhances downstream tasks, low-level imaging metrics are employed. Specifically, counting inlier matches and measuring the inlier-outlier ratio over time. Using such metrics helps asserting tracking, mapping, or recognition tasks are possible. The results are compared in Fig.~\ref{fig:inliers} and Table~\ref{tab:inlier} where in all cases, the image processing pipeline is the same. First, contrast limited adaptive histogram equalisation (CLAHE) \cite{pizer_adaptive_1987} is applied, and scale-invariant feature transform (SIFT) \cite{lowe_distinctive_2004} features are extracted with a reduced contrast threshold. Feature matching is then performed on consecutive frames, and outlier rejection is done using random sample consensus (RANSAC) \cite{fischler_random_1981}. On average, the proposed approach enables a higher ratio and number of inliers in low turbidity water and a higher ratio in highly turbid water. Additionally, the results are more stable over depth as the method naturally maintains a number of matches. The NIQE approach performs well in high turbidity and is able to maintain a significant number of inliers, however, in low turbidity, because the normalised local statistics of the images do not present significant changes, it favours remaining at a high range, and thus losing mapping quality as the vehicle gets deeper. In comparison, the gradient based approach is highly unstable. This is because it is dependent on the quantity of features present in the image, which can bias the results if the scene is inconsistent. This led to average performance in both water conditions. In clear water, the pattern with 1m distance performs best at the first depth because of its proximity to the target and the excellent light condition, however, at the third and fourth depths, it is not performing as well because at these depths, more artificial light is required to obtain more contrast. The pattern at 2m distance is getting good results in clear water until the environment gets too dark because of the depth. However, in the second environment, the water is too turbid to be able to extract enough features at a 2m distance.

\begin{table}[t]
\caption{Visual coverage area evaluation in different turbidity levels\label{tab:coverage}}
\centering
\begin{tabular}{ccc}
\toprule
Setup                  & Low ($m^2$) & High ($m^2$) \\ \midrule
Proposed                        & 35.9                                    & 16.5                                     \\
NIQE                            & \textbf{56.3}                           & 13.5                                     \\
Gradient                        & 33.1                                    & 16.1                                     \\
1m pattern                      & 15.8                                    & 15.8                                     \\
2m pattern                      & 39.4                                    & \textbf{19.7}                            \\
\bottomrule
\end{tabular}
\end{table}

Finally, the visual coverage is considered. It is estimated based on the field of view (FOV) of the camera and the distance to the target. An area is considered visually covered only if its features can be extracted. Table~\ref{tab:coverage} displays the resulting coverage. All the methods perform significantly better than the 1m pattern in clear water. The 2m pattern and the NIQE approach tend to cover a larger area, but at the cost of quality, which is significantly degraded. In high turbidity, both the gradient method and the 2m pattern ends up not being able to extract usable features at some steps of the operation.

Overall, the developed approach proposes the best trade-off between inspection quality and coverage. Compared to the other approaches, it also offers stability making possible efficient operations. Furthermore, its capability to adapt to local illumination changes and water column properties shows the model can generalise well.

\section{Limitations}

The model relies on the diversity of data for efficient training. Although the model used during the experiments was trained on a very large variety of water types, only relatively shallow depths and two surface textures were considered, requiring the model to extrapolate in deep waters and drastically different scenery, which might lead to increased prediction errors. Similarly for the distance to target, far distances were not considered.

Furthermore, the calibration procedure requires the vehicle to possess high manoeuvrability, precise station keeping, or a tilting camera, as the vehicle is required to look both towards the sea surface and horizontally. Horizontal movements should be minimised while estimating the depth profile, and similarly, in the vertical plane when estimating the contrast profiles.

The framework might fail in the presence of structurally complex underwater assets, especially if the scene depth has a large variation, as in this work the inspected assets were considered as relatively planar, e.g., a ship hull or a smooth seabed. This is because a single value is used to determine the distance to the target. Considering multiple points and spatially distributed would help overcome this issue.

\section{Conclusion}

In this article, an active perception framework enabling underwater robotic operations in challenging conditions was presented. It enables deployments in conditions that were previously prohibitive. To achieve this, the system was designed to understand the water column properties and provide suggestions to the vehicle to obtain better quality imagery. The approach was successfully tested and compared in simulations, including turbid and clear water environments.

Future work includes field trials and further generalising the method by developing a more complete model of the water column, including parameters such as density of suspended particulate matter and how it affects active illumination. Additionally, defining backscattering in the model as a function of depth and light intensity would help guide the vehicle more precisely.

\section*{Acknowledgments}
This work was supported by the ARC Research Hub in Intelligent Robotic Systems for Real-Time Asset Management (IH210100030). The authors would like to thank Peter Roberts and Julien Flack from Advanced Navigation for their support and contribution to this work.

\bibliographystyle{IEEEtran}
\bibliography{references.bib}

\end{document}